\documentclass[11pt,a4paper]{article}
\usepackage[hyperref]{naaclhlt2019}
\usepackage{times}
\usepackage{latexsym}
\usepackage{multicol}
\usepackage{amsfonts}
\usepackage{comment}
\usepackage{tikz}
\usepackage{pgfplots}
\usetikzlibrary{shapes, positioning, decorations.pathreplacing, shapes.multipart}
\usepackage{amsmath}
\usepackage{todonotes}
\usepackage{algorithm}
\usepackage{multirow}
\usepackage{subfig}
\usepackage[noend]{algpseudocode}
\usepackage{booktabs}
\usepackage{url}

\aclfinalcopy 

\def\bs{\mathbf{s}}
\def\bh{\mathbf{h}}
\def\be{\mathbf{e}}
\DeclareMathOperator*{\Softmax}{\mathrm{softmax}}
\DeclareMathOperator*{\Att}{Att}

\title{Star-Transformer}

\author{
  Qipeng Guo\footnotemark[2] \\qpguo16@fudan.edu.cn \\\And Xipeng Qiu\footnotemark[2]\hspace{1mm}  \thanks{\hspace{1mm}  Corresponding Author.} \\xpqiu@fudan.edu.cn \\\And Pengfei Liu\footnotemark[2] \\pfliu14@fudan.edu.cn \\\AND Yunfan Shao\footnotemark[2] \\yfshao15@fudan.edu.cn \\\And Xiangyang Xue\footnotemark[2] \\xyxue@fudan.edu.cn\\\\
\footnotemark[2]\hspace{0.5mm}  Shanghai Key Laboratory of Intelligent Information Processing, Fudan University\\
\footnotemark[2]\hspace{0.5mm} School of Computer Science, Fudan University\\
\setcounter{footnote}{2}
\footnotemark[3]\hspace{0.5mm} New York University \\\And Zheng Zhang\thanks{\hspace{1mm} work done at NYU Shanghai, now with AWS Shanghai AI Lab.}  \\zz@nyu.edu
}

\date{}

\begin{document}
\maketitle
\begin{abstract}


Although \textit{Transformer} has achieved great successes on many NLP tasks, its heavy structure with fully-connected attention connections leads to dependencies on large training data.
In this paper, we present \textit{Star-Transformer}, a lightweight alternative by careful sparsification. 
To reduce model complexity, we replace the fully-connected structure with a star-shaped topology, in which every two non-adjacent nodes are connected through a shared relay node. Thus, complexity is reduced from quadratic to linear, while preserving the capacity to capture both local composition and long-range dependency. 
The experiments on four tasks (22 datasets) show that Star-Transformer achieved significant improvements against the standard Transformer for the modestly sized datasets.
\end{abstract}

\section{Introduction}

Recently, the fully-connected attention-based models, like Transformer \cite{DBLP:conf/nips/VaswaniSPUJGKP17}, become popular in natural language processing (NLP) applications, notably machine translation \cite{DBLP:conf/nips/VaswaniSPUJGKP17} and language modeling \cite{radford2018improving}. Some recent work also suggest that Transformer can be an alternative to recurrent neural networks (RNNs) and convolutional neural networks (CNNs) in many NLP tasks, such as GPT \cite{radford2018improving}, BERT \cite{DBLP:journals/corr/abs-1810-04805}, Transformer-XL \cite{DBLP:journals/corr/abs-1901-02860} and Universal Transformer \cite{DBLP:journals/corr/abs-1807-03819}. 

\begin{figure}
\subfloat{}{
\resizebox {0.45\linewidth} {!} {
\begin{tikzpicture}[font=\LARGE\selectfont]
\foreach \a in {1,2,...,8}{
\draw[thick] (-\a*360/8+180-360/8: 3cm) node[circle,draw=black!60](\a){$\mathbf{h}_\a$};
}
\draw[draw=black,thick,-](1)--(2)--(3)--(4)--(5)--(6)--(7)--(8)--(1);
\draw[draw=black,thick,-](1)--(3)--(5)--(7);
\draw[draw=black,thick,-](1)--(4)--(6)--(8)--(2);
\draw[draw=black,thick,-](1)--(5)--(8)--(3);
\draw[draw=black,thick,-](1)--(6)--(2)--(4)--(7);
\draw[draw=black,thick,-](1)--(7)--(2)--(5);
\draw[draw=black,thick,-](8)--(4);
\draw[draw=black,thick,-](6)--(3)--(7);
\end{tikzpicture}
}
}
\hfill
\subfloat{}{
\resizebox {0.45\linewidth} {!} {
\begin{tikzpicture}[font=\LARGE\selectfont]
\node[rectangle,draw=black!60,minimum size=10mm](0,0)(C){$\mathbf{s}$};
\foreach \a in {1,2,...,8}{
\draw[thick] (-\a*360/8+180-360/8: 3cm) node[circle,draw=black!60](\a){$\mathbf{h}_\a$};
\draw[draw=blue,thick,-](\a)--(C);
}
\draw[draw=red,thick,-](1)--(2)--(3)--(4)--(5)--(6)--(7)--(8)--(1);
\end{tikzpicture}
}
}
\caption{Left: Connections of one layer in Transformer, circle nodes indicate the hidden states of input tokens. Right: Connections of one layer in Star-Transformer, the square node is the virtual relay node. Red edges and blue edges are ring and radial connections, respectively.}
\label{fig:main}
\end{figure}

More specifically, there are two limitations of the Transformer. First, the computation and memory overhead of the Transformer are quadratic to the sequence length. This is especially problematic with long sentences. Transformer-XL \cite{DBLP:journals/corr/abs-1901-02860} provides a solution which achieves the acceleration and performance improvement, but it is specifically designed for the language modeling task. Second, studies indicate that Transformer would fail on many tasks if the training data is limited, unless it is pre-trained on a large corpus. \cite{radford2018improving,DBLP:journals/corr/abs-1810-04805}.


A key observation is that Transformer does not exploit prior knowledge well. For example, the local compositionality is already a robust inductive bias for modeling the text sequence. However, the Transformer learns this bias from scratch, along with non-local compositionality, thereby increasing the learning cost. The key insight is then whether leveraging strong prior knowledge can help to ``lighten up'' the architecture.

To address the above limitation, we proposed a new lightweight model named ``Star-Transformer''. The core idea is to sparsify the architecture by moving the fully-connected topology into a star-shaped structure. Fig-\ref{fig:main} gives an overview. Star-Transformer has two kinds of connections. The radial connections preserve the non-local communication and remove the redundancy in fully-connected network. The ring connections embody the local-compositionality prior, which has the same role as in CNNs/RNNs. The direct outcome of our design is the improvement of both efficiency and learning cost: the computation cost is reduced from quadratic to linear as a function of input sequence length. An inherent advantage is that the ring connections can effectively reduce the burden of the unbias learning of local and non-local compositionality and improve the generalization ability of the model. What remains to be tested is whether one shared relay node is capable of capturing the long-range dependencies. 

We evaluate the Star-Transformer on three NLP tasks including Text Classification, Natural Language Inference, and Sequence Labelling. Experimental results show that Star-Transformer outperforms the standard Transformer consistently and has less computation complexity. An additional analysis on a simulation task indicates that Star-Transformer preserve the ability to handle with long-range dependencies which is a crucial feature of the standard Transformer.

In this paper, we claim three contributions as the following and our code is available on Github \footnote{\url{https://github.com/dmlc/dgl} and \url{https://github.com/fastnlp/fastNLP}}:
\begin{itemize}
    \item Compared to the standard Transformer, Star-Transformer has a lightweight structure but with an approximate ability to model the long-range dependencies. It reduces the number of connections from $n^2$ to $2n$, where $n$ is the sequence length.
    \item
         The Star-Transformer divides the labor of semantic compositions between the radial and the ring connections. The radial connections focus on the non-local compositions and the ring connections focus on the local composition. Therefore, Star-Transformer works for modestly sized datasets and does not rely on heavy pre-training.
    \item We design a simulation task ``Masked Summation'' to probe the ability dealing with long-range dependencies. In this task, we verify that both Transformer and Star-Transformer are good at handling long-range dependencies compared to the LSTM and BiLSTM.
\end{itemize}

\section{Related Work}
Recently, neural networks have proved very successful in learning text representation and have achieved state-of-the-art results in many different tasks.

\paragraph{Modelling Local Compositionality}

A popular approach is to represent each word as a low-dimensional vector and then learn the local semantic composition functions over the given sentence structures. For example,
\citet{kim2014convolutional,kalchbrenner2014convolutional} used CNNs to capture the semantic representation of sentences, whereas \citet{cho2014learning} used RNNs.

These methods are biased for learning local compositional functions and are hard to capture the long-term dependencies in a text sequence. In order to augment the ability to model the non-local compositionality, a class of improved methods utilizes various self-attention mechanisms to aggregate the weighted information of each word, which can be used to get sentence-level representations for classification tasks \cite{yang2016hierarchical,lin2017structured,DBLP:conf/aaai/ShenZLJPZ18}.
Another class of improved methods augments neural networks with a re-reading ability or global state while processing each word \cite{cheng2016long,DBLP:conf/acl/ZhangLS18}.

\paragraph{Modelling Non-Local Compositionality}

There are two kinds of methods to model the non-local semantic compositions in a text sequence directly.

One class of models incorporate syntactic tree into the network structure for learning sentence representations \cite{DBLP:conf/acl/TaiSM15,zhu2015long}.

Another type of models learns the dependencies between words based entirely on self-attention without any recurrent or convolutional layers, such as Transformer \cite{DBLP:conf/nips/VaswaniSPUJGKP17}, which has achieved state-of-the-art results on a machine translation task.
The success of Transformer has raised a large body of follow-up work. Therefore, some Transformer variations are also proposed, such as GPT \cite{radford2018improving},  BERT \cite{DBLP:journals/corr/abs-1810-04805}, Transformer-XL \cite{DBLP:journals/corr/abs-1901-02860} , Universal Transformer \cite{DBLP:journals/corr/abs-1807-03819} and CN$^3$ \cite{liu2018contextualized}.


However, those Transformer-based methods usually require a large training corpus. When applying them on modestly sized datasets, we need the help of semi-supervised learning and unsupervised pretraining techniques \cite{radford2018improving}.

\paragraph{Graph Neural Networks}
Star-Transformer is also inspired by the recent graph networks \cite{gilmer2017neural, kipf2016semi,battaglia2018relational,liu2018multi}, in which the information fusion progresses via message-passing across the whole graph.

The graph structure of the Star-Transformer is star-shaped by introducing a virtual relay node. The radial and ring connections give a better balance between the local and non-local compositionality.
Compared to the previous augmented models \cite{yang2016hierarchical,lin2017structured,DBLP:conf/aaai/ShenZLJPZ18,cheng2016long,DBLP:conf/acl/ZhangLS18}, the implementation of Star-Transform is purely based on the attention mechanism similar to the standard Transformer, which is simpler and well suited for parallel computation.

Due to its better parallel capacity and lower complexity, the Star-Transformer is faster than RNNs or Transformer, especially on modeling long sequences.


\section{Model}

\subsection{Architecture}

The Star-Transformer consists of one relay node and $n$ satellite nodes. The state of $i$-th satellite node represents the features of the $i$-th token in a text sequence. The relay node acts as a virtual hub to gather and scatter information from and to all the satellite nodes.

Star-Transformer has a star-shaped structure, with two kinds of connections in the: the radial connections and the ring connections.

\paragraph{Radial Connections}
For a network of $n$ satellite nodes, there are $n$ radial connections. Each connection links a satellite node to the shared relay node.  With the radial connections, every two non-adjacent satellite nodes are two-hop neighbors and can receive non-local information with a two-step update.

\paragraph{Ring Connections}
Since text input is a sequence, we bake such prior as an inductive bias. Therefore, we connect the adjacent satellite nodes to capture the relationship of local compositions. The first and last nodes are also connected. Thus, all these local connections constitute a ring-shaped structure.
Note that the ring connections allow each satellite node to gather information from its neighbors and plays the same role to CNNs or bidirectional RNNs.

With the radial and ring connections, Star-Transformer can capture both the non-local and local compositions simultaneously.
Different from the standard Transformer, we make a division of labor, where the radial connections capture non-local compositions, whereas the ring connections attend to local compositions.

\subsection{Implementation}

The implementation of the Star-Transformer is very similar to the standard Transformer, in which the information exchange is based on the attention mechanism \cite{DBLP:conf/nips/VaswaniSPUJGKP17}.


\paragraph{Multi-head Attention}

Just as in the standard Transformer, we use the scaled dot-product attention \cite{DBLP:conf/nips/VaswaniSPUJGKP17}. Given a sequence of vectors $\mathbf{H} \in \mathbb{R}^{n\times d}$, we can use a query vector $\mathbf{q}\in \mathbb{R}^{1\times d} $ to soft select the relevant information with attention.
\begin{align}
    \Att(\mathbf{q},\mathbf{K},\mathbf{V}) = \Softmax(\frac{\mathbf{q}\mathbf{K}^T}{\sqrt{d}})\mathbf{V},
\end{align}
where $\mathbf{K} = \mathbf{H} \mathbf{W}^K,\mathbf{V}=\mathbf{H} \mathbf{W}^V$, and $\mathbf{W}^K, \mathbf{W}^V$ are learnable parameters.

To gather more useful information from $\mathbf{H}$, similar to multi-channels in CNNs, we can use multi-head attention with $k$ heads.
\begin{align}
    &\mathrm{MultiAtt}(\mathbf{q},\mathbf{H}) = (\mathbf{a}_1 \oplus\cdots\oplus \mathbf{a}_k)\mathbf{W}^O , \\
    & \mathbf{a}_i = \Att(\mathbf{q}\mathbf{W}_i^Q, \mathbf{H}\mathbf{W}_i^K, \mathbf{H}\mathbf{W}_i^V), i\in[1,k]
\end{align}
where $\oplus$ denotes the concatenation operation, and  $\mathbf{W}_i^Q, \mathbf{W}_i^K, \mathbf{W}_i^V, \mathbf{W}^O$ are learnable parameters.

\paragraph{Update}

Let $\mathbf{s}^t  \in \mathbb{R}^{1\times d}$ and $\mathbf{H}^t \in \mathbb{R}^{n\times d}$ denote the states for the relay node and all the $n$ satellite nodes at step $t$.
When using the Star-Transformer to encode a text sequence of length $n$, we start from its embedding $\mathbf{E}=[\be_1;\cdots; \be_n]$, where $\be_i \in \mathbb{R}^{1\times d}$ is the embedding of the i-th token.

We initialize the state with $\mathbf{H}^0 =\mathbf{E}$ and $\mathbf{s}^0=average(\mathbf{E})$.

The update of the Star-Transformer at step $t$ can be divided into two alternative phases: (1) the update of the satellite nodes and (2) the update of the relay node.

At the first phase, the state of each satellite node $\mathbf{h}_i$ are updated from its adjacent nodes, including the neighbor nodes $\mathbf{h}_{i-1},\mathbf{h}_{i+1}$ in the sequence, the relay node $\bs^t$, its previous state, and its corresponding token embedding.
\begin{align}
    \mathbf{C}^t_i &= [\bh^{t-1}_{i-1}; \bh^{t-1}_{i}; \bh^{t-1}_{i+1}; \be^i; \bs^{t-1}], \\
    \bh^t_i &= \mathrm{MultiAtt}(\bh^{t-1}_i, \mathbf{C}^t_i ),
\end{align}
where $\mathbf{C}^t_i$ denotes the context information for the $i$-th satellite node. Thus, the update of each satellite node is similar to the recurrent network, except that the update fashion is based on attention mechanism.
After the information exchange, a layer normalization operation \cite{DBLP:journals/corr/BaKH16} is used.
\begin{align}
    \bh^t_i &= \mathrm{LayerNorm}(\mathrm{ReLU}(\bh^t_i)), i\in[1,n].
\end{align}

At the second phase, the relay node $\bs^t$ summarizes the information of all the satellite nodes and its previous state.
\begin{align}
   \bs^t &= \mathrm{MultiAtt}(\bs^{t-1}, [\bs^{t-1};\mathbf{H}^t]),\\
   \bs^t &= \mathrm{LayerNorm}(\mathrm{ReLU}(\bs^t)).
\end{align}

By alternatively updating update the satellite and relay nodes, the Star-Transformer finally captures all the local and non-local compositions for an input text sequence.

\paragraph{Position Embeddings}
To incorporate the sequence information,
we also add the learnable position embeddings, which are added with the token embeddings at the first layer.

The overall update algorithm of the Star-Transformer is shown in the Alg-\ref{alg:main}. 

\begin{algorithm}[t]

  \caption{The Update of Star-Transformer}
  \label{alg:main}
  Input: Number of layers $T$, embedding of input tokens $e_1, \cdots, e_n$
  \begin{algorithmic}[1]
  \State  \textcolor[rgb]{0.00,0.59,0.00}{{// \textit{Initialization}}}
  \State  $\bh^0_1,\cdots, \bh^0_n \leftarrow \be_1,\cdots , \be_n$
  \State $\bs^0 \leftarrow average(\be_1,\cdots, \be_n)$

  \For{$t$ \textbf{from} $1$ \textbf{to} $T$}
    \State  \textcolor[rgb]{0.00,0.59,0.00}{{// \textit{update the satellite nodes}}}
    \For{$i$ \textbf{from} $1$ \textbf{to} $n$}
     \State $\mathbf{C}^t_i = [\bh^{t-1}_{i-1};\bh^{t-1}_{i}; \bh^{t-1}_{i+1}; \be^i;\bs^{t-1}]$
     \State $\bh^t_i = \mathrm{MultiAtt}(\bh^{t-1}_i, \mathbf{C}^t_i)$
    \State  $\bh^t_i = \mathrm{LayerNorm}(\mathrm{ReLU}(\bh^t_i))$

    \EndFor
    \State  \textcolor[rgb]{0.00,0.59,0.00}{{// \textit{update the relay node}}}
    \State $\bs^t = \mathrm{MultiAtt}(\bs^{t-1}, [\bs^{t-1};\mathbf{H}^t])$
    \State $\bs^t = \mathrm{LayerNorm}(\mathrm{ReLU}(\bs^t))$
  \EndFor
  \end{algorithmic}
\end{algorithm}

\subsection{Output}
After $T$ rounds of update, the final states of $\mathbf{H}^T$ and $\mathbf{s}^T$ can be used for various tasks such as sequence labeling and classification. For different tasks, we feed them to different task-specific modules.
For classification, we generate the fix-length sentence-level vector representation by applying a max-pooling across the final layer and mixing it with $\mathbf{s}^T$, this vector is fed into a Multiple Layer Perceptron (MLP) classifier. For the sequence labeling task, the $\mathbf{H}^T$  provides features corresponding to all the input tokens.

\begin{table*}[t]
    \centering
    \small
    \begin{tabular}{c|c*7{c}}
    \toprule
    \multicolumn{2}{c}{Dataset} & Train & Dev. & Test & $|V|$ & H DIM & \#head & head DIM \\
    \midrule
    \multicolumn{2}{c}{Masked Summation} & 10k &10k& 10k & - & 100 & 10 & 10 \\
    \midrule
    \multicolumn{2}{c}{SST \cite{DBLP:conf/emnlp/SocherPWCMNP13}} & 8k &1k& 2k & 20k & 300 & 6 & 50\\
    \midrule
    \multirow{5}{1.5cm}{MTL-16 $^\dagger$ \cite{DBLP:conf/acl/LiuQH17}} & \multirow{5}{4cm}{{\small Apparel Baby Books Camera DVD Electronics Health IMDB Kitchen Magazines MR Music Software Sports Toys Video} } & \multirow{5}{*}{1400} & \multirow{5}{*}{200} & \multirow{5}{*}{400} & \multirow{5}{*}{8k$\sim$28k} & \multirow{5}{*}{300} & \multirow{5}{*}{6} & \multirow{5}{*}{50} \\
    & & & & & & & &\\
    & & & & & & & &\\
    & & & & & & & &\\
    & & & & & & & &\\
    \midrule
    \multicolumn{2}{c}{SNLI \cite{DBLP:conf/emnlp/BowmanAPM15}} & 550k &10k & 10k & 34k & 600 & 6 & 100 \\
    \midrule
    \multicolumn{2}{c}{PTB POS \cite{DBLP:journals/coling/MarcusSM94}} & 38k &5k& 5k & 44k & 300 & 6 & 50 \\
    \midrule
    \multicolumn{2}{c}{CoNLL03 \cite{DBLP:conf/conll/SangM03}} &15k & 3k& 3k & 25k & 300 & 6 & 50 \\
    \midrule
    \multicolumn{2}{c}{OntoNotes NER \cite{DBLP:conf/conll/PradhanMXUZ12}} &94k &14k& 8k & 63k & 300 & 6 & 50 \\
    \bottomrule
    \end{tabular}
    \caption{An overall of datasets and its hyper-parameters, ``H DIM, \#head, head DIM" indicates the dimension of hidden states, the number of heads in the Multi-head attention, the dimension of each head, respectively. MTL-16$^\dagger$ consists of 16 datasets, each of them has 1400/200/400 samples in train/dev/test.  }
    \label{tab:exp_all}
\end{table*}

\section{Comparison to the standard Transformer}

Since our goal is making the Transformer lightweight and easy to train with modestly sized dataset, we have removed many connections compared with the standard Transformer (see Fig-\ref{fig:main}). If the sequence length is $n$ and the dimension of hidden states is $d$, the computation complexity of one layer in the standard Transformer is $O(n^2 d)$. The Star-Transformer has two phases, the update of ring connections costs $O(5nd )$ (the constant $5$ comes from the size of context information $\mathbf{C}$), and the update of radial connections costs $O(nd)$, so the total cost of one layer in the Star-Transformer is $O(6nd)$.

In theory, Star-Transformer can cover all the possible relationships in the standard Transformer. For example, any relationship $\bh_i \rightarrow \bh_j$ in the standard Transformer can be simulated by $\bh_i \rightarrow \bs \rightarrow \bh_j$. The experiment on the simulation task in Sec-\ref{sec:toy_task} provides some evidence to show the virtual node $\bs$ could handle long-range dependencies. Following this aspect, we can give a rough analysis of the path length of dependencies in these models. As discussed in the Transformer paper \cite{DBLP:conf/nips/VaswaniSPUJGKP17}, the maximum dependency path length of RNN and Transformer are $O(n)$, $O(1)$, respectively. Star-Transformer can pass the message from one node to another node via the relay node so that the maximum dependency path length is also $O(1)$, with a constant two comparing to Transformer.

Compare with the standard Transformer, all positions are processed in parallel, pair-wise connections are replaced with a  ``gather and dispatch'' mechanism. As a result, we accelerate the Transformer 10 times on the simulation task and 4.5 times on real tasks. The model also preserves the ability to handle long input sequences. Besides the acceleration, the Star-Transformer achieves significant improvement on some modestly sized datasets.

\tikzset{
  halfnode/.style={inner sep=.5em, draw, rectangle split, rectangle split parts=#1},
  innernode/.style={inner sep=.3333em, draw, rectangle split, rectangle split parts=#1}
}
\begin{figure*}
    \centering
    \pgfplotsset{width=0.8\textwidth}
    \begin{tikzpicture}[font=\small\selectfont]
    \node[rectangle, draw=white] (0) {Input};
    \node[innernode=3, draw=blue] (1) [below=0.2 of 0]  {1 \nodepart{second} 0.3 \nodepart{third} 0.4 };
    \node[innernode=3, draw=black] (2) [right=of 1]  {0 \nodepart{second} 0.5 \nodepart{third} 0.7 };
    \node[innernode=3, draw=black] (3) [right=of 2] {0 \nodepart{second} 0.1 \nodepart{third} 0.2 };
    \node[innernode=3, draw=blue] (4) [right=of 3]  {1 \nodepart{second} 0.5 \nodepart{third} 0.9 };
    \node[innernode=3, draw=black] (5) [right=of 4] {0 \nodepart{second} 0.4 \nodepart{third} 0.2 };
    \node[innernode=3, draw=black] (6) [right=of 5] {0 \nodepart{second} 0.6 \nodepart{third} 0.8 };
    \node[innernode=3, draw=black] (7) [right=of 6] {0 \nodepart{second} 0.1 \nodepart{third} 0.3 };
    \node[innernode=3, draw=blue] (8) [right=of 7] {1 \nodepart{second} 0.1 \nodepart{third} 0.6 };
    \draw[red,-,thick] (12.5, -2)--(12.5, 0);
    \node[halfnode=2, draw=black] (10)[right= 1.0 of 8] { 0.9 \nodepart{second} 1.9};    \node[rectangle, draw=white] (11) [above=0.2 of 10] {Target};
    \end{tikzpicture}
    \caption{An example of the masked summation task, the input is a sequence of $n$ vectors, each vector has $d$ dimension, and there are total $k$ vectors which have the mask value equals $1$. The target is the summation of such masked vectors. In this figure, $n=8, k=3, d=3$.}
    \label{fig:toy_case}
\end{figure*}

\section{Experiments}

We evaluate Star-Transformer on one simulation task to probe its behavior when challenged with long-range dependency problem, and three real tasks (Text Classification, Natural Language Inference, and Sequence Labelling). All experiments are ran on a NVIDIA Titan X card. Datasets used in this paper are listed in the Tab-\ref{tab:exp_all}. We use the Adam \cite{DBLP:journals/corr/KingmaB14} as our optimizer. On the real task, we set the embedding size to 300 and initialized with GloVe \cite{DBLP:conf/emnlp/PenningtonSM14}. And the symbol ``Ours + Char'' means an additional character-level pre-trained embedding JMT \cite{DBLP:conf/emnlp/HashimotoXTS17} is used. Therefore, the total size of embedding should be 400 which as a result of the concatenation of GloVe and JMT. We also fix the embedding layer of the Star-Transformer in all experiments.

Since semi- or unsupervised model is also a feasible solution to improve the model in a parallel direction, such as the ELMo \cite{DBLP:conf/naacl/PetersNIGCLZ18} and BERT \cite{DBLP:journals/corr/abs-1810-04805}, we exclude these models in the comparison and focus on the relevant architectures.

\subsection{Masked Summation}
\label{sec:toy_task}
In this section, we introduce a simulation task on the synthetic data to probe the efficiency and non-local/long-range dependencies of LSTM, Transformer, and the Star-Transformer. As mentioned in \cite{DBLP:conf/nips/VaswaniSPUJGKP17}, the maximum path length of long-range dependencies of LSTM and Transformer are $O(n)$ and $O(1)$, where $n$ is the sequence length. The maximum dependency path length of Star-Transformer is $O(1)$ with a constant two via the relay node. To validate the ability to deal with long-range dependencies, we design a simulation task named ``Masked Summation''. The input of this task is a matrix $\mathbf{X} \in \mathbb{R}^{n \times d}$, it has $n$ columns and each column has $d$ elements. The first dimension indicates the mask value $\mathbf{X}_{i0} \in \{0, 1\}$, $0$ means the column is ignored in summation. The rest $d-1$ elements are real numbers in drawn uniformly from the range $[0, 1)$. The target is a $d-1$ dimensional vector which equals the summation of all the columns with the mask value $1$. There is an implicit variable $k$ to control the number of $1$ in the input. Note that a simple baseline is always guessing the value $k/2$.

The evaluation metric is the Mean Square Error (MSE), and the generated dataset has (10k/10k/10k) samples in (train/dev/test) sets. The Fig-\ref{fig:toy_case} show a case of the masked summation task.

\begin{figure}[t]
    \subfloat[MSE loss on the masked summation when $n=200,k=10,d=10$. The $k/2$ line means the MSE loss when the model always guess the exception value $k/2$.]{
    \includegraphics[width=0.8\linewidth]{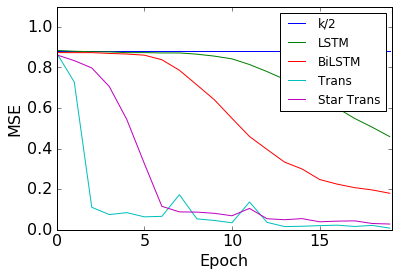}
    \label{fig:toy_exp1}
    }
    \hfill
    \subfloat[Test Time, $k=10,d=10$]{
    \includegraphics[width=0.8\linewidth]{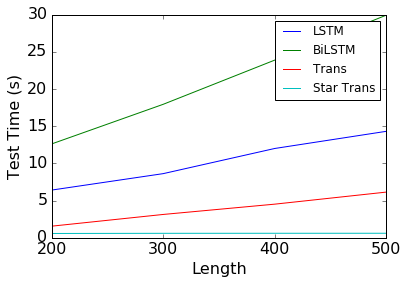}
    \label{fig:toy_exp2}
    }
    \label{fig:my_label}
\end{figure}

The mask summation task asks the model to recognize the mask value and gather columns in different positions. When the sequence length $n$ is significantly higher than the number of the columns $k$, the model will face the long-range dependencies problem. The Fig-\ref{fig:toy_exp1} shows the performance curves of models on various lengths. Although the task is easy, the performance of LSTM and BiLSTM dropped quickly when the sequence length increased. However, both Transformer and Star-Transformer performed consistently on various lengths. The result indicates the Star-Transformer preserves the ability to deal with the non-local/long-range dependencies.

Besides the performance comparison, we also study the speed with this simulation task since we could ignore the affection of padding, masking, and data processing. We also report the inference time in the Fig-\ref{fig:toy_exp2}, which shows that Transformer is faster than LSTM and BiLSTM a lot, and Star-Transformer is faster than Transformer, especially on the long sequence.

\subsection{Text Classification}
Text classification is a basic NLP task, and we select two datasets to observe the performance of our model in different conditions, Stanford Sentiment Treebank(SST) dataset \cite{DBLP:conf/emnlp/SocherPWCMNP13} and MTL-16 \cite{DBLP:conf/acl/LiuQH17} consists of 16 small datasets on various domains. We truncate the sequence which its length higher than 256 to ensure the standard Transformer can run on a single GPU card.

\begin{table}
    \centering\small
    \begin{tabular}{lc}
        \toprule
        Model & Acc \\
        \midrule
        BiLSTM \cite{DBLP:conf/emnlp/LiLJH15} & 49.8  \\

        Tree-LSTM \cite{DBLP:conf/acl/TaiSM15}  & 51.0 \\

        CNN-Tensor \cite{DBLP:conf/emnlp/LeiBJ15}  & 51.2 \\

        Emb + self-att \cite{DBLP:conf/aaai/ShenZLJPZ18}  & 48.9 \\

        BiLSTM + self-att \cite{DBLP:journals/corr/abs-1808-07383}  & 50.4 \\

        CNN + self-att \cite{DBLP:journals/corr/abs-1808-07383} & 50.6  \\
        Dynamic self-att \cite{DBLP:journals/corr/abs-1808-07383}  & 50.6 \\

        DiSAN \cite{DBLP:conf/aaai/ShenZLJPZ18}& 51.7  \\

        \midrule
        Transformer & 50.4 \\

        Ours &  \textbf{52.9}   \\
        \bottomrule
    \end{tabular}
    \caption{Test Accuracy on SST dataset.}
    \label{tab:exp_cls1}
\end{table}

\begin{table*}
    \centering\small\setlength{\tabcolsep}{3pt}
    \begin{tabular}{l*4{c}|*4{c}}
        \toprule
        \bf \multirow{2}{*}{Dataset} & \multicolumn{4}{c|}{Acc (\%)} & \multicolumn{3}{c}{Test Time (ms)} & \multirow{2}{*}{Len.} \\
        & Star-Transformer & Transformer & BiLSTM & SLSTM & Star-Transformer & Transformer & BiLSTM & \\
        \midrule
        Apparel & 88.25 & 82.25 & 86.05 & 85.75 & 11 & 34 &  114 & 65 \\
        
        Baby & 87.75 & 84.50 & 84.51 & 86.25 & 11 & 50 & 141 & 109 \\
        
        Books & 87.00 & 81.50 & 82.12 & 83.44 & 11 & 55 & 135 & 131 \\
       
        Camera & 92.25 & 86.00 & 87.05 & 90.02 & 11 & 52 & 125 & 116 \\
        
        DVD & 86.00 & 77.75 & 83.71 & 85.52 & 11 & 58 & 143 & 139 \\
        
        Electronics & 82.75 & 81.50 & 82.51 & 83.25 & 10 & 49 & 136 & 105 \\
        
        Health & 87.50 & 83.50 & 85.52 & 86.50 & 10 & 45 & 131 & 85 \\
       
        IMDB & 88.00 & 82.50 & 86.02 & 87.15 & 12 & 77 & 174 & 201 \\
        
        Kitchen & 85.50 & 83.00 & 82.22 & 84.54 & 11 & 47 & 130 & 92 \\
        
        Magazines & 92.75 & 89.50 & 92.52 & 93.75 & 11 & 51 & 136 & 115 \\
        
        MR & 79.00 & 77.25 & 75.73 & 76.20 & 12 & 14 & 26 & 22 \\
        
        Music & 83.50 & 79.00 & 78.74 & 82.04 & 11 & 53 & 137 & 118 \\
        
        Software & 90.50 & 85.25 & 86.73 & 87.75 & 11 & 53 & 140 & 117 \\
        
        Sports & 86.25 & 84.75 & 84.04 & 85.75 & 10 & 49 & 134 & 103 \\
        
        Toys & 88.00 & 82.00 &  85.72 & 85.25 & 11 & 46 & 137 & 96 \\
        
        Video & 86.75 & 84.25 & 84.73 & 86.75 & 11 & 56 & 146 & 131 \\
        
        \midrule
        \bf Average & \textbf{86.98} & 82.78 & 84.01 & 85.38  & \textbf{10.94}  & 49.31 & 130.3 & 109.1  \\
        \bottomrule
    \end{tabular}
    \caption{Test Accuracy over MTL-16 datasets. ``Test Time" means millisecond per batch on the test set (batch\_size is 128). ``Len.'' means the average sequence length on the test set.}
    \label{tab:exp_cls2}
\end{table*}

For classification tasks, we use the state of the relay node $\mathbf{s}^T$ plus the feature of max pooling on satellite nodes $\max(\mathbf{H}^T)$ as the final representation and feed it into the softmax classifier. The description of hyper-parameters is listed in Tab-\ref{tab:exp_all} and Appendix.

Results on SST and MTL-16 datasets are listed in Tab-\ref{tab:exp_cls1},\ref{tab:exp_cls2}, respectively.
On the SST, the Star-Transformer achieves 2.5 points improvement against the standard Transformer and beat the most models.

Also, on the MTL-16, the Star-Transformer outperform the standard Transformer in all 16 datasets, the improvement of the average accuracy is 4.2. The Star-Transformer also gets better results compared with existing works. As we mentioned in the introduction, the standard Transformer requires large training set to reveal its power. Our experiments show the Star-Transformer could work well on the small dataset which only has 1400 training samples. Results of the time-consuming show the Star-Transformer could be 4.5 times fast than the standard Transformer on average.

\subsection{Natural Language Inference}
Natural Language Inference (NLI) asks the model to identify the semantic relationship between a premise sentence and a corresponding hypothesis sentence. In this paper, we use the Stanford Natural Language Inference (SNLI) \cite{DBLP:conf/emnlp/BowmanAPM15} for evaluation. Since we want to study how the model encodes the sentence as a vector representation, we set Star-Transformer as a sentence vector-based model and compared it with sentence vector-based models.

In this experiment, we follow the previous work \cite{DBLP:conf/acl/BowmanGRGMP16} to use $\text{concat}( \mathbf{r}_1, \mathbf{r}_2, \| \mathbf{r}_1 - \mathbf{r}_2 \|, \mathbf{r}_1 - \mathbf{r}_2) $ as the classification feature. The $ \mathbf{r}_1, \mathbf{r}_2$ are representations of premise and hypothesis sentence, it is calculated by $\mathbf{s}^T +\max(\mathbf{H}^T) $ which is same with the classification task. See Appendix for the detail of hyper-parameters.

\begin{table}[t]
    \centering\small
    \begin{tabular}{p{6cm}c}
    \toprule
        Model & Acc\\
    \midrule
        BiLSTM \cite{DBLP:journals/corr/LiuSLW16} & 83.3  \\
        BiLSTM + self-att \cite{DBLP:journals/corr/LiuSLW16} & 84.2   \\

        300D SPINN-PI \cite{DBLP:conf/acl/BowmanGRGMP16} & 83.2  \\

        Tree-based CNN \cite{DBLP:conf/acl/MouMLX0YJ16} & 82.1 \\

        4096D BiLSTM-max \cite{DBLP:conf/emnlp/ConneauKSBB17} & 84.5   \\
        300D DiSAN \cite{DBLP:conf/aaai/ShenZLJPZ18} & 85.6  \\
        Residual encoders \cite{DBLP:conf/repeval/NieB17} & 86.0 \\
        Gumbel TreeLSTM \cite{DBLP:conf/aaai/ChoiYL18} & 86.0  \\

        Reinforced self-att \cite{DBLP:conf/ijcai/ShenZLJWZ18} & 86.3 \\
        2400D Multiple DSA \cite{DBLP:journals/corr/abs-1808-07383} & 87.4  \\
        \midrule

        Transformer & 82.2 \\

        Star-Transformer & \textbf{86.0} \\
    \bottomrule
    \end{tabular}
    \caption{Test Accuracy on SNLI dataset. }
    \label{tab:nli_exp}
\end{table}

\begin{table*}[t]
    \centering\small
    \begin{tabular}{p{5cm}|*2{c}|c|*2{c}}
    \toprule
        \multirow{3}{*}{Model} & \multicolumn{2}{|c|}{\multirow{2}{*}{Adv Tech}} & POS & \multicolumn{2}{|c}{NER} \\
        \cline{4-6}
        & \multicolumn{2}{|c|}{} &  PTB  & CoNLL2003 & CoNLL2012 \\
        \cline{2-6}
        &char & CRF  &  Acc & F1 & F1 \\
        \midrule
        \cite{DBLP:conf/emnlp/LingDBTFAML15} &\checkmark & \checkmark  & 97.78  & - & - \\

        \cite{DBLP:journals/jmlr/CollobertWBKKK11}  &\checkmark & \checkmark  & 97.29  & 89.59 & - \\

        \cite{DBLP:journals/corr/HuangXY15} &\checkmark & \checkmark  & 97.55 & 90.10 & - \\

        \cite{chiu2016sequential} &\checkmark &\checkmark   &  -  & 90.69 & 86.35 \\

        \cite{DBLP:conf/acl/MaH16} &\checkmark & \checkmark  & 97.55 &  91.06 & - \\

        \cite{DBLP:journals/tacl/NguyenTW16} &\checkmark & \checkmark  & -& 91.2 & - \\

        \cite{DBLP:journals/tacl/ChiuN16} &\checkmark & \checkmark & - & 91.62 & 86.28 \\

        \cite{DBLP:conf/acl/ZhangLS18}& \checkmark & \checkmark  & 97.55  & 91.57 & - \\

        \cite{DBLP:journals/corr/abs-1808-03926} &\checkmark & \checkmark & 97.43  & 91.11 & 87.84 \\

        \midrule
        Transformer& & &  96.31  & 86.48 & 83.57 \\

        Transformer + Char& \checkmark & &  97.04  & 88.26 & 85.14 \\

        Star-Transformer& & &  97.14  & 90.93 & 86.30 \\

        Star-Transformer + Char& \checkmark & & 97.64 & 91.89 & 87.64\\

        Star-Transformer + Char + CRF& \checkmark & \checkmark  & \textbf{97.68} & \textbf{91.98} & \textbf{87.88} \\
        \bottomrule
    \end{tabular}
    \caption{Results on sequence labeling tasks. We list the ``Advanced Techniques'' except widely-used pre-trained embeddings (GloVe, Word2Vec, JMT) in columns. The ``Char" indicates character-level features, it also includes the Capitalization Features, Suffix Feature, Lexicon Features, etc. The ``CRF" means an additional Conditional Random Field (CRF) layer.}
    \label{tab:seq_exp}
\end{table*}
As shown in Tab-\ref{tab:nli_exp}, the Star-Transformer outperforms most typical baselines (DiSAN, SPINN) and achieves comparable results compared with the state-of-the-art model.
Notably, our model beats standard Transformer by a large margin, which is easy to overfit although we have made a careful hyper-parameters' searching for Transformer.

The SNLI dataset is not a small dataset in NLP area, so improving the generalization ability of the Transformer is a significant topic.

The best result in Tab-\ref{tab:nli_exp} \cite{DBLP:journals/corr/abs-1808-07383} using a large network and fine-tuned hyper-parameters, they get the best result on SNLI but an undistinguished result on SST, see Tab-\ref{tab:exp_cls1}.

\subsection{Sequence Labelling}
To verify the ability of our model in sequence labeling, we choose two classic sequence labeling tasks: Part-of-Speech  (POS) tagging and Named Entity Recognition (NER) task.

Three datasets are used as our benchmark: one POS tagging dataset from Penn Treebank (PTB) \cite{DBLP:journals/coling/MarcusSM94}, and two NER datasets from CoNLL2003 \cite{DBLP:conf/conll/SangM03}, CoNLL2012 \cite{DBLP:conf/conll/PradhanMXUZ12}. We use the final state of satellite nodes $\mathbf{H}^T$ to classify the label in each position. Since we believe that the complex neural network could be an alternative of the CRF, we also report the result without CRF layer.

As shown in Tab-\ref{tab:seq_exp}, Star-Transformer achieves the state-of-the-art performance on sequence labeling tasks. The ``Star-Transformer + Char" has already beat most of the competitors.
Star-Transformer could achieve such results without CRF, suggesting that the model has enough capability to capture the partial ability of the CRF. The Star-Transformer also outperforms the standard Transformer on sequence labeling tasks with a significant gap.  
\subsection{Ablation Study}
\begin{table}[t]
    \centering\small
    \begin{tabular}{lccc}
        \toprule
        \multirow{2}{*}{Model} & SNLI & CoNLL03 & MS \\
        & Acc & Acc & MSE \\
        \midrule
        Star-Transformer & 86.0  & 90.93 & 0.0284\\

        \hspace{1em} variant (a) -radial & 84.0 & 89.35  & 0.1536\\

        \hspace{1em} variant (b) -ring & 77.6  & 79.36  & 0.0359\\
        \bottomrule
    \end{tabular}
    \caption{Test Accuracy on SNLI dataset, CoNLL2003 NER task and the Masked Summation $n=200,k=10,d=10$.  }
    \label{tab:ablation}
\end{table}
In this section, we perform an ablation study to test the effectiveness of the radial and ring connections.

We test two variants of our models, the first variants (a) remove the radial connections and only keep the ring connections. Without the radial connections, the maximum path length of this variant becomes $O(n)$. The second variant (b) removes the ring connections and remains the radial connections.
Results in Tab-\ref{tab:ablation} give some insights, the variant (a) loses the ability to handle long-range dependencies, so it performs worse on both the simulation and real tasks. However, the performance drops on SNLI and CoNLL03 is moderate since the remained ring connections still capture the local features. The variant (b) still works on the simulation task since the maximum path length stays unchanged. Without the ring connections, it loses its performance heavily on real tasks. Therefore, both the radial and ring connections are necessary to our model.

\section{Conclusion and Future Works}
In this paper, we present Star-Transformer which reduce the computation complexity of the standard Transformer by carefully sparsifying the topology. We compare the standard Transformer with other models on one toy dataset and 21 real datasets and find Star-Transformer outperforms the standard Transformer and achieves comparable results with state-of-the-art models.

This work verifies the ability of Star-Transformer by excluding the factor of unsupervised pre-training. In the future work, we will investigate the ability of Star-Transformer by unsupervised pre-training on the large corpus. Moreover, we also want to introduce more NLP prior knowledge into the model.

\section*{Acknowledgments}
We would like to thank the anonymous reviewers for their valuable comments. The research work is supported by Shanghai Municipal Science and Technology Commission (No. 17JC1404100 and 16JC1420401),
National Key Research and Development Program of China (No. 2017YFB1002104),
and National Natural Science Foundation of China (No. 61672162 and 61751201).

\bibliography{naaclhlt2019,nlp,nlp1,ours}

\begin{thebibliography}{49}
\expandafter\ifx\csname natexlab\endcsname\relax\def\natexlab#1{#1}\fi

\bibitem[{Akhundov et~al.(2018)Akhundov, Trautmann, and
  Groh}]{DBLP:journals/corr/abs-1808-03926}
Adnan Akhundov, Dietrich Trautmann, and Georg Groh. 2018.
\newblock Sequence labeling: {A} practical approach.
\newblock \emph{CoRR}, abs/1808.03926.

\bibitem[{Ba et~al.(2016)Ba, Kiros, and Hinton}]{DBLP:journals/corr/BaKH16}
Lei~Jimmy Ba, Ryan Kiros, and Geoffrey~E. Hinton. 2016.
\newblock Layer normalization.
\newblock \emph{CoRR}, abs/1607.06450.

\bibitem[{Battaglia et~al.(2018)Battaglia, Hamrick, Bapst, Sanchez-Gonzalez,
  Zambaldi, Malinowski, Tacchetti, Raposo, Santoro, Faulkner
  et~al.}]{battaglia2018relational}
Peter~W Battaglia, Jessica~B Hamrick, Victor Bapst, Alvaro Sanchez-Gonzalez,
  Vinicius Zambaldi, Mateusz Malinowski, Andrea Tacchetti, David Raposo, Adam
  Santoro, Ryan Faulkner, et~al. 2018.
\newblock Relational inductive biases, deep learning, and graph networks.
\newblock \emph{arXiv preprint arXiv:1806.01261}.

\bibitem[{Bowman et~al.(2015)Bowman, Angeli, Potts, and
  Manning}]{DBLP:conf/emnlp/BowmanAPM15}
Samuel~R. Bowman, Gabor Angeli, Christopher Potts, and Christopher~D. Manning.
  2015.
\newblock A large annotated corpus for learning natural language inference.
\newblock In \emph{{EMNLP}}, pages 632--642. The Association for Computational
  Linguistics.

\bibitem[{Bowman et~al.(2016)Bowman, Gauthier, Rastogi, Gupta, Manning, and
  Potts}]{DBLP:conf/acl/BowmanGRGMP16}
Samuel~R. Bowman, Jon Gauthier, Abhinav Rastogi, Raghav Gupta, Christopher~D.
  Manning, and Christopher Potts. 2016.
\newblock A fast unified model for parsing and sentence understanding.
\newblock In \emph{{ACL} {(1)}}. The Association for Computer Linguistics.

\bibitem[{Cheng et~al.(2016)Cheng, Dong, and Lapata}]{cheng2016long}
Jianpeng Cheng, Li~Dong, and Mirella Lapata. 2016.
\newblock Long short-term memory-networks for machine reading.
\newblock In \emph{Proceedings of the 2016 Conference on EMNLP}, pages
  551--561.

\bibitem[{Chiu and Nichols(2016{\natexlab{a}})}]{chiu2016sequential}
Jason Chiu and Eric Nichols. 2016{\natexlab{a}}.
\newblock Sequential labeling with bidirectional lstm-cnns.
\newblock In \emph{Proc. International Conf. of Japanese Association for NLP},
  pages 937--940.

\bibitem[{Chiu and Nichols(2016{\natexlab{b}})}]{DBLP:journals/tacl/ChiuN16}
Jason P.~C. Chiu and Eric Nichols. 2016{\natexlab{b}}.
\newblock Named entity recognition with bidirectional lstm-cnns.
\newblock \emph{{TACL}}, 4:357--370.

\bibitem[{Cho et~al.(2014)Cho, van Merrienboer, Gulcehre, Bougares, Schwenk,
  and Bengio}]{cho2014learning}
Kyunghyun Cho, Bart van Merrienboer, Caglar Gulcehre, Fethi Bougares, Holger
  Schwenk, and Yoshua Bengio. 2014.
\newblock Learning phrase representations using rnn encoder-decoder for
  statistical machine translation.
\newblock In \emph{Proceedings of EMNLP}.

\bibitem[{Choi et~al.(2018)Choi, Yoo, and Lee}]{DBLP:conf/aaai/ChoiYL18}
Jihun Choi, Kang~Min Yoo, and Sang{-}goo Lee. 2018.
\newblock Learning to compose task-specific tree structures.
\newblock In \emph{{AAAI}}, pages 5094--5101. {AAAI} Press.

\bibitem[{Collobert et~al.(2011)Collobert, Weston, Bottou, Karlen, Kavukcuoglu,
  and Kuksa}]{DBLP:journals/jmlr/CollobertWBKKK11}
Ronan Collobert, Jason Weston, L{\'{e}}on Bottou, Michael Karlen, Koray
  Kavukcuoglu, and Pavel~P. Kuksa. 2011.
\newblock Natural language processing (almost) from scratch.
\newblock \emph{Journal of Machine Learning Research}, 12:2493--2537.

\bibitem[{Conneau et~al.(2017)Conneau, Kiela, Schwenk, Barrault, and
  Bordes}]{DBLP:conf/emnlp/ConneauKSBB17}
Alexis Conneau, Douwe Kiela, Holger Schwenk, Lo{\"{\i}}c Barrault, and Antoine
  Bordes. 2017.
\newblock Supervised learning of universal sentence representations from
  natural language inference data.
\newblock In \emph{{EMNLP}}, pages 670--680. Association for Computational
  Linguistics.

\bibitem[{Dai et~al.(2019)Dai, Yang, Yang, Carbonell, Le, and
  Salakhutdinov}]{DBLP:journals/corr/abs-1901-02860}
Zihang Dai, Zhilin Yang, Yiming Yang, Jaime~G. Carbonell, Quoc~V. Le, and
  Ruslan Salakhutdinov. 2019.
\newblock Transformer-xl: Attentive language models beyond a fixed-length
  context.
\newblock \emph{CoRR}, abs/1901.02860.

\bibitem[{Dehghani et~al.(2018)Dehghani, Gouws, Vinyals, Uszkoreit, and
  Kaiser}]{DBLP:journals/corr/abs-1807-03819}
Mostafa Dehghani, Stephan Gouws, Oriol Vinyals, Jakob Uszkoreit, and Lukasz
  Kaiser. 2018.
\newblock Universal transformers.
\newblock \emph{CoRR}, abs/1807.03819.

\bibitem[{Devlin et~al.(2018)Devlin, Chang, Lee, and
  Toutanova}]{DBLP:journals/corr/abs-1810-04805}
Jacob Devlin, Ming{-}Wei Chang, Kenton Lee, and Kristina Toutanova. 2018.
\newblock {BERT:} pre-training of deep bidirectional transformers for language
  understanding.
\newblock \emph{CoRR}, abs/1810.04805.

\bibitem[{Gilmer et~al.(2017)Gilmer, Schoenholz, Riley, Vinyals, and
  Dahl}]{gilmer2017neural}
Justin Gilmer, Samuel~S Schoenholz, Patrick~F Riley, Oriol Vinyals, and
  George~E Dahl. 2017.
\newblock Neural message passing for quantum chemistry.
\newblock In \emph{ICML}, pages 1263--1272.

\bibitem[{Hashimoto et~al.(2017)Hashimoto, Xiong, Tsuruoka, and
  Socher}]{DBLP:conf/emnlp/HashimotoXTS17}
Kazuma Hashimoto, Caiming Xiong, Yoshimasa Tsuruoka, and Richard Socher. 2017.
\newblock A joint many-task model: Growing a neural network for multiple {NLP}
  tasks.
\newblock In \emph{{EMNLP}}, pages 1923--1933. Association for Computational
  Linguistics.

\bibitem[{Huang et~al.(2015)Huang, Xu, and Yu}]{DBLP:journals/corr/HuangXY15}
Zhiheng Huang, Wei Xu, and Kai Yu. 2015.
\newblock Bidirectional {LSTM-CRF} models for sequence tagging.
\newblock \emph{CoRR}, abs/1508.01991.

\bibitem[{Kalchbrenner et~al.(2014)Kalchbrenner, Grefenstette, and
  Blunsom}]{kalchbrenner2014convolutional}
Nal Kalchbrenner, Edward Grefenstette, and Phil Blunsom. 2014.
\newblock A convolutional neural network for modelling sentences.
\newblock In \emph{Proceedings of ACL}.

\bibitem[{Kim(2014)}]{kim2014convolutional}
Yoon Kim. 2014.
\newblock Convolutional neural networks for sentence classification.
\newblock In \emph{Proceedings of the 2014 Conference on EMNLP}, pages
  1746--1751.

\bibitem[{Kingma and Ba(2014)}]{DBLP:journals/corr/KingmaB14}
Diederik~P. Kingma and Jimmy Ba. 2014.
\newblock Adam: {A} method for stochastic optimization.
\newblock \emph{CoRR}, abs/1412.6980.

\bibitem[{Kipf and Welling(2016)}]{kipf2016semi}
Thomas~N Kipf and Max Welling. 2016.
\newblock Semi-supervised classification with graph convolutional networks.
\newblock \emph{arXiv preprint arXiv:1609.02907}.

\bibitem[{Lei et~al.(2015)Lei, Barzilay, and
  Jaakkola}]{DBLP:conf/emnlp/LeiBJ15}
Tao Lei, Regina Barzilay, and Tommi~S. Jaakkola. 2015.
\newblock Molding cnns for text: non-linear, non-consecutive convolutions.
\newblock In \emph{{EMNLP}}, pages 1565--1575. The Association for
  Computational Linguistics.

\bibitem[{Li et~al.(2015)Li, Luong, Jurafsky, and
  Hovy}]{DBLP:conf/emnlp/LiLJH15}
Jiwei Li, Thang Luong, Dan Jurafsky, and Eduard~H. Hovy. 2015.
\newblock When are tree structures necessary for deep learning of
  representations?
\newblock In \emph{{EMNLP}}, pages 2304--2314. The Association for
  Computational Linguistics.

\bibitem[{Lin et~al.(2017)Lin, Feng, Xiang, Zhou, and
  Bengio}]{lin2017structured}
Zhouhan Lin, Mo~Feng, Yu, Bing Xiang, Bowen Zhou, and Yoshua Bengio. 2017.
\newblock A structured self-attentive sentence embedding.
\newblock \emph{arXiv preprint arXiv:1703.03130}.

\bibitem[{Ling et~al.(2015)Ling, Dyer, Black, Trancoso, Fermandez, Amir,
  Marujo, and Lu{\'{\i}}s}]{DBLP:conf/emnlp/LingDBTFAML15}
Wang Ling, Chris Dyer, Alan~W. Black, Isabel Trancoso, Ramon Fermandez, Silvio
  Amir, Lu{\'{\i}}s Marujo, and Tiago Lu{\'{\i}}s. 2015.
\newblock Finding function in form: Compositional character models for open
  vocabulary word representation.
\newblock In \emph{{EMNLP}}, pages 1520--1530. The Association for
  Computational Linguistics.

\bibitem[{Liu et~al.(2018{\natexlab{a}})Liu, Chang, Huang, Tang, and
  Cheung}]{liu2018contextualized}
Pengfei Liu, Shuaichen Chang, Xuanjing Huang, Jian Tang, and Jackie Chi~Kit
  Cheung. 2018{\natexlab{a}}.
\newblock Contextualized non-local neural networks for sequence learning.
\newblock \emph{arXiv preprint arXiv:1811.08600}.

\bibitem[{Liu et~al.(2018{\natexlab{b}})Liu, Fu, Dong, Qiu, and
  Cheung}]{liu2018multi}
Pengfei Liu, Jie Fu, Yue Dong, Xipeng Qiu, and Jackie Chi~Kit Cheung.
  2018{\natexlab{b}}.
\newblock Multi-task learning over graph structures.
\newblock \emph{arXiv preprint arXiv:1811.10211}.

\bibitem[{Liu et~al.(2017)Liu, Qiu, and Huang}]{DBLP:conf/acl/LiuQH17}
Pengfei Liu, Xipeng Qiu, and Xuanjing Huang. 2017.
\newblock Adversarial multi-task learning for text classification.
\newblock In \emph{{ACL} {(1)}}, pages 1--10. Association for Computational
  Linguistics.

\bibitem[{Liu et~al.(2016)Liu, Sun, Lin, and
  Wang}]{DBLP:journals/corr/LiuSLW16}
Yang Liu, Chengjie Sun, Lei Lin, and Xiaolong Wang. 2016.
\newblock Learning natural language inference using bidirectional {LSTM} model
  and inner-attention.
\newblock \emph{CoRR}, abs/1605.09090.

\bibitem[{Ma and Hovy(2016)}]{DBLP:conf/acl/MaH16}
Xuezhe Ma and Eduard~H. Hovy. 2016.
\newblock End-to-end sequence labeling via bi-directional lstm-cnns-crf.
\newblock In \emph{{ACL} {(1)}}. The Association for Computer Linguistics.

\bibitem[{Marcus et~al.(1993)Marcus, Santorini, and
  Marcinkiewicz}]{DBLP:journals/coling/MarcusSM94}
Mitchell~P. Marcus, Beatrice Santorini, and Mary~Ann Marcinkiewicz. 1993.
\newblock Building a large annotated corpus of english: The penn treebank.
\newblock \emph{Computational Linguistics}, 19(2):313--330.

\bibitem[{Mou et~al.(2016)Mou, Men, Li, Xu, Zhang, Yan, and
  Jin}]{DBLP:conf/acl/MouMLX0YJ16}
Lili Mou, Rui Men, Ge~Li, Yan Xu, Lu~Zhang, Rui Yan, and Zhi Jin. 2016.
\newblock Natural language inference by tree-based convolution and heuristic
  matching.
\newblock In \emph{{ACL} {(2)}}. The Association for Computer Linguistics.

\bibitem[{Nguyen et~al.(2016)Nguyen, Theobald, and
  Weikum}]{DBLP:journals/tacl/NguyenTW16}
Dat~Ba Nguyen, Martin Theobald, and Gerhard Weikum. 2016.
\newblock {J-NERD:} joint named entity recognition and disambiguation with rich
  linguistic features.
\newblock \emph{{TACL}}, 4:215--229.

\bibitem[{Nie and Bansal(2017)}]{DBLP:conf/repeval/NieB17}
Yixin Nie and Mohit Bansal. 2017.
\newblock Shortcut-stacked sentence encoders for multi-domain inference.
\newblock In \emph{RepEval@EMNLP}, pages 41--45. Association for Computational
  Linguistics.

\bibitem[{Pennington et~al.(2014)Pennington, Socher, and
  Manning}]{DBLP:conf/emnlp/PenningtonSM14}
Jeffrey Pennington, Richard Socher, and Christopher~D. Manning. 2014.
\newblock Glove: Global vectors for word representation.
\newblock In \emph{{EMNLP}}, pages 1532--1543. {ACL}.

\bibitem[{Peters et~al.(2018)Peters, Neumann, Iyyer, Gardner, Clark, Lee, and
  Zettlemoyer}]{DBLP:conf/naacl/PetersNIGCLZ18}
Matthew~E. Peters, Mark Neumann, Mohit Iyyer, Matt Gardner, Christopher Clark,
  Kenton Lee, and Luke Zettlemoyer. 2018.
\newblock Deep contextualized word representations.
\newblock In \emph{{NAACL-HLT}}, pages 2227--2237. Association for
  Computational Linguistics.

\bibitem[{Pradhan et~al.(2012)Pradhan, Moschitti, Xue, Uryupina, and
  Zhang}]{DBLP:conf/conll/PradhanMXUZ12}
Sameer Pradhan, Alessandro Moschitti, Nianwen Xue, Olga Uryupina, and Yuchen
  Zhang. 2012.
\newblock Conll-2012 shared task: Modeling multilingual unrestricted
  coreference in ontonotes.
\newblock In \emph{EMNLP-CoNLL Shared Task}, pages 1--40. {ACL}.

\bibitem[{Radford et~al.(2018)Radford, Narasimhan, Salimans, and
  Sutskever}]{radford2018improving}
Alec Radford, Karthik Narasimhan, Tim Salimans, and Ilya Sutskever. 2018.
\newblock Improving language understanding by generative pre-training.

\bibitem[{Sang and Meulder(2003)}]{DBLP:conf/conll/SangM03}
Erik F. Tjong~Kim Sang and Fien~De Meulder. 2003.
\newblock Introduction to the conll-2003 shared task: Language-independent
  named entity recognition.
\newblock In \emph{CoNLL}, pages 142--147. {ACL}.

\bibitem[{Shen et~al.(2018{\natexlab{a}})Shen, Zhou, Long, Jiang, Pan, and
  Zhang}]{DBLP:conf/aaai/ShenZLJPZ18}
Tao Shen, Tianyi Zhou, Guodong Long, Jing Jiang, Shirui Pan, and Chengqi Zhang.
  2018{\natexlab{a}}.
\newblock Disan: Directional self-attention network for rnn/cnn-free language
  understanding.
\newblock In \emph{{AAAI}}, pages 5446--5455. {AAAI} Press.

\bibitem[{Shen et~al.(2018{\natexlab{b}})Shen, Zhou, Long, Jiang, Wang, and
  Zhang}]{DBLP:conf/ijcai/ShenZLJWZ18}
Tao Shen, Tianyi Zhou, Guodong Long, Jing Jiang, Sen Wang, and Chengqi Zhang.
  2018{\natexlab{b}}.
\newblock Reinforced self-attention network: a hybrid of hard and soft
  attention for sequence modeling.
\newblock In \emph{{IJCAI}}, pages 4345--4352. ijcai.org.

\bibitem[{Socher et~al.(2013)Socher, Perelygin, Wu, Chuang, Manning, Ng, and
  Potts}]{DBLP:conf/emnlp/SocherPWCMNP13}
Richard Socher, Alex Perelygin, Jean Wu, Jason Chuang, Christopher~D. Manning,
  Andrew~Y. Ng, and Christopher Potts. 2013.
\newblock Recursive deep models for semantic compositionality over a sentiment
  treebank.
\newblock In \emph{{EMNLP}}, pages 1631--1642. {ACL}.

\bibitem[{Tai et~al.(2015)Tai, Socher, and Manning}]{DBLP:conf/acl/TaiSM15}
Kai~Sheng Tai, Richard Socher, and Christopher~D. Manning. 2015.
\newblock Improved semantic representations from tree-structured long
  short-term memory networks.
\newblock In \emph{{ACL} {(1)}}, pages 1556--1566. The Association for Computer
  Linguistics.

\bibitem[{Vaswani et~al.(2017)Vaswani, Shazeer, Parmar, Uszkoreit, Jones,
  Gomez, Kaiser, and Polosukhin}]{DBLP:conf/nips/VaswaniSPUJGKP17}
Ashish Vaswani, Noam Shazeer, Niki Parmar, Jakob Uszkoreit, Llion Jones,
  Aidan~N. Gomez, Lukasz Kaiser, and Illia Polosukhin. 2017.
\newblock Attention is all you need.
\newblock In \emph{{NIPS}}, pages 6000--6010.

\bibitem[{Yang et~al.(2016)Yang, Yang, Dyer, He, Smola, and
  Hovy}]{yang2016hierarchical}
Zichao Yang, Diyi Yang, Chris Dyer, Xiaodong He, Alex Smola, and Eduard Hovy.
  2016.
\newblock Hierarchical attention networks for document classification.
\newblock In \emph{Proceedings of the 2016 Conference of NAACL}, pages
  1480--1489.

\bibitem[{Yoon et~al.(2018)Yoon, Lee, and
  Lee}]{DBLP:journals/corr/abs-1808-07383}
Deunsol Yoon, Dongbok Lee, and SangKeun Lee. 2018.
\newblock Dynamic self-attention : Computing attention over words dynamically
  for sentence embedding.
\newblock \emph{CoRR}, abs/1808.07383.

\bibitem[{Zhang et~al.(2018)Zhang, Liu, and Song}]{DBLP:conf/acl/ZhangLS18}
Yue Zhang, Qi~Liu, and Linfeng Song. 2018.
\newblock Sentence-state {LSTM} for text representation.
\newblock In \emph{{ACL} {(1)}}, pages 317--327. Association for Computational
  Linguistics.

\bibitem[{Zhu et~al.(2015)Zhu, Sobhani, and Guo}]{zhu2015long}
Xiao-Dan Zhu, Parinaz Sobhani, and Hongyu Guo. 2015.
\newblock Long short-term memory over recursive structures.
\newblock In \emph{ICML}, pages 1604--1612.

\end{thebibliography}
\bibliographystyle{acl_natbib}

\end{document}